\documentclass[11pt, a4paper, singlecolumn]{article}

\usepackage{color}
\usepackage{ulem} 

\usepackage[utf8]{inputenc}
\usepackage[english]{babel}
\usepackage{comment}
\usepackage{booktabs}

\usepackage[top=2cm, bottom=2cm, left=1.9cm, right=1.9cm]{geometry}

\usepackage{titlesec}
\titlespacing{\section}{0pt}{\parskip}{\parskip}
\titlespacing{\subsection}{0pt}{\parskip}{\parskip}
\titlespacing{\subsubsection}{0pt}{\parskip}{\parskip}

\usepackage{indentfirst}
\usepackage[export]{adjustbox}
\usepackage{graphicx}
\usepackage{amssymb}
\usepackage{amsmath}

\usepackage[dvipsnames]{xcolor}

\usepackage[figurename=Fig.]{caption}
\usepackage{caption}
\usepackage[colorlinks, citecolor=ForestGreen]{hyperref}

\usepackage{cite}

\usepackage{abstract}
\usepackage{multirow}
\usepackage{array}
\newcolumntype{*}{>{\global\let\currentrowstyle\relax}}
\newcolumntype{^}{>{\currentrowstyle}}

\title{\textbf{Adversarial Sparse Teacher: Defense Against Distillation-Based Model Stealing Attacks Using Adversarial Examples}}
\date{\vspace*{2pt}}
\author{
	\normalsize \textbf{Eda Yilmaz}\\
	\normalsize Hacettepe University\\
	\normalsize Computer Engineering Department\\
	\normalsize yilmaz-eda@hacettepe.edu.tr
	\and
	\normalsize \textbf{Hacer Yalim Keles}\\
	\normalsize Hacettepe University\\
	\normalsize Computer Engineering Department\\
	\normalsize hacerkeles@cs.hacettepe.edu.tr
}

\begin{document}
\maketitle
\begin{abstract}
We introduce Adversarial Sparse Teacher (AST), a robust defense method against distillation-based model stealing attacks. Our approach trains a teacher model using adversarial examples to produce sparse logit responses and increase the entropy of the output distribution. Typically, a model generates a peak in its output corresponding to its prediction. By leveraging adversarial examples, AST modifies the teacher model's original response, embedding a few altered logits into the output while keeping the primary response slightly higher. Concurrently, all remaining logits are elevated to further increase the output distribution's entropy. All these complex manipulations are performed using an optimization function with our proposed Exponential Predictive Divergence (EPD) loss function. EPD allows us to maintain higher entropy levels compared to traditional KL divergence, effectively confusing attackers. Experiments on CIFAR-10 and CIFAR-100 datasets demonstrate that AST outperforms state-of-the-art methods, providing effective defense against model stealing while preserving high accuracy. The source codes will be made publicly available \href{https://github.com/codeofanon/AdversarialSparseTeacher/}{here} soon.

\end{abstract}
\section{Introduction}
\label{sec:intro}
In deep learning, the efficiency and performance of neural networks are highly important,particularly in applications where computational resources are limited.Knowledge Distillation(KD) introduces a method in this context, enabling the knowledge transfer from a complex, high-performing teacher model to a more simpler, efficient student model \cite{KD}.The realm of research surrounding this concept has significantly expanded. The methodology of KD can be broadly categorized based on the type of the knowledge transferred: output responses\cite{KD,mirzadeh2020improved}, features\cite{romero2015fitnets,chen2021feature,komodakis2017paying}, and relations\cite{park2019relational,passalis2020heterogeneous,peng2019correlation}. Each category targets a different aspect of the teacher model's intelligence; from its output predictions to the intermediate representations and the inter-layer interactions or relationship between samples, respectively\cite{gou2021knowledge}.An interesting technique of KD is Self Distillation,where a model is trained to distill knowledge from itself, effectively refining its performance without the need for an external teacher model\cite{SD,yuan2020revisiting}. 

Knowledge distillation facilitates the deployment of powerful neural networks on resource limited devices and enhances the performance of smaller models. However, these methods has increased concerns regarding security of proprietary models through model stealing attacks. Such attacks aim to replicate the functionality  of proprietary models without authorized access, posing significant threats to intellectual property and model integrity\cite{IP}. Various methods have been proposed to affirm ownership of machine learning models, including watermarking\cite{adi2018turning,uchida2017embedding} and passport-based protections\cite{fan2019rethinking,zhang2020passport}. However, these methods remained insufficient in preventing the extraction of models.

The potential theft of valuable models necessitates proactive measures for prevention. A study called \textit{Nasty Teacher} addresses this issue\cite{NT}. This method involves creating a secondary model, derived from the base model, which is resistant to model stealing while maintaining comparable performance. The approach involves subtly altering the output logits of the secondary model, which effectively misleads any adversarial \textit{student} models attempting to replicate it, without altering the primary behavior or performance of the model under standard conditions. To achieve this, the algorithm increases the relative entropy between the output distributions of the base and secondary models, utilizing Kullback-Leibler (KL) divergence\cite{KL}, while simultaneously preserving high accuracy through minimization of cross-entropy loss.

An extension of this research, known as the \textit{Stingy Teacher}, takes a different approach by directly inducing sparsity in the output logits. This method structures sparse logits by zeroing out class probabilities except for the top $n$ classes prior to knowledge distillation, revealing only this reduced information\cite{ST}. This approach, theoretically validated to fail student models, inspired our method's incorporation of sparsity. Like \textit{Nasty Teacher}, our method aligns with the principle of maintaining a cautious teacher, but instead of applying modifications to the outputs, we integrate sparsity into the solution. This not only enhances the undistillability of the teacher but also addresses the challenges associated with creating a teacher model whose sparse outputs remain immutable in the face of distillation attempts.

Inspired by these studies, our work represents a novel defensive method called \textit{Adversarial Sparse Teacher (AST)}. In contrast to the \textit{Nasty Teacher} model, which advances learning by concurrently maximizing Kullback-Leibler (KL) divergence and minimizing cross-entropy loss, our approach adopts a contrary strategy. We prioritize \textit{minimizing} all the terms in the loss, aiming to steer learning in a specific, well-defined direction. This approach ensures a more balanced path in optimization. Moreover, we propose and use a new divergence function, \textit{Exponential Predictive Divergence (EPD)} that eliminates the unfavorable effects of KL divergence during optimization.Our method employs \textit{adversarial counterparts of the original images} to systematically generate ambiguous outputs relative to the given inputs. The non-random nature of these adversarial images ensures a consistent redundancy in the output logits, effectively deceiving potential model stealers.

The creation of adversarial examples is crucial for understanding their nature and impact on neural networks. 
Projected Gradient Descent (PGD) method, begins at a random point and iteratively applies perturbations until reaching the maximum allowed magnitude\cite{PGD}.

Furthermore, to enhance defense against adversarial attacks, defense strategies often incorporate adversarial examples into model training. This can involve training models directly on these examples or pairing the outputs from adversarial examples with those from clean examples. The Adversarial Logit Pairing (ALP) method\cite{ALP} exemplifies this strategy by showing that minimizing the L2 distance between the logits of clean images and their adversarial counterparts mitigates adversarial attacks. This indicates that the features of adversarial images can be learned and countered through the analysis of logits.

Building upon these insights, our proposed teacher model includes a sparsity constraint, aiming to encapsulate adversarial information that maximizes ambiguity within its logits while preserving high accuracy. To achieve this, we defined a new loss function that distills from a sparse representation of the model’s response to adversarial images. To the best of our knowledge, this approach is novel in this domain. Our work makes two significant contributions to the field, summarized as follows:
\begin{itemize}
    \item We introduce \textit{Adversarial Sparse Teacher} (AST), a novel methodology that leverages sparse outputs of adversarial examples as targets during training to mislead the student with a novel loss function. This approach is designed to enhance the robustness of models against stealing attacks, addressing a critical concern in contemporary model security.
    \item We propose a new \textit{Exponential Divergence Function} (EPD), which provides a novel method for assessing discrepancies between predicted and actual probability distributions using an exponential coefficient. Our experiments demonstrate that EPD is particularly effective in maintaining sparse peaks within a high-entropy setting. This allows the defensive model to retain its original predictions within an ambiguous logit distribution, thereby confusing the student model while preserving the performance of the teacher model.
\end{itemize}

\section{Methodology}
\subsection{Knowledge Distillation}
In the context of knowledge distillation, the objective is to train a student neural network, which is denoted as $f_{\theta _S}( \cdot )$, to mimic the behavior of a pre-trained teacher network, denoted as $f_{\theta _T}( \cdot )$. Both networks have their respective parameter sets: ${\theta _T}$  for the teacher network and ${\theta _S}$ for the student network. The training dataset contains pairs of samples and their corresponding labels, denoted as \((x^{(i)}, y^{(i)})\). The logit response of a sample $x^{(i)}$ from the network $f_{\theta }( \cdot )$ is indicated by $p_{f_{\theta }}( x^{(i)} )$. The \textit{softmax temperature} function, $\sigma_{\tau_s}(\cdot) $, which is proposed by Hinton et al. in 2015, transforms logits into soft probabilities when a large temperature ${\tau_s}$, which is usually greater than 1, is applied; and it behaves like the standard softmax function ${\sigma}(\cdot)$, when ${\tau_S}$ is set to 1 \cite{KD}. To achieve knowledge distillation, a combined loss function is utilized for training the student network, as shown in Eqn. (\ref{eq:example}).
\begin{align}
 \mathcal{L_{KD}}=
    \alpha \tau_S^2 \; \mathcal{L_{KL}}(\sigma_{\tau_s}(p_{f_{\theta_T}}(x^{(i)})), \sigma_{\tau_s}(p_{f_{\theta_S}}(x^{(i)}))) \nonumber \\ + (1-\alpha) \mathcal{L_{CE}}(\sigma(p_{f_{\theta_S}}(x^{(i)})), y^{(i)})\label{eq:example}
\end{align}
 This combined loss function consists of two terms. The first term measures the KL divergence between the softened logits of the teacher network, $\sigma_{\tau_S}(p_{f_{\theta _T}}( x^{(i)} ))$, and the student network, $\sigma_{\tau_S}(p_{f_{\theta _S}}( x^{(i)} ))$. This divergence quantifies the difference in the distributions of soft probabilities generated by the two networks. The second term is the cross-entropy loss between the softened probabilities of the student network, $\sigma {(p_{f_{\theta _S}}( x^{(i)} ))}$, and the ground truth label, $y^{(i)}$. This term ensures that the student network is learning to predict the correct labels directly. 

To control the balance between knowledge distillation and conventional cost minimization, a hyper-parameter $\alpha$ is usually used. By adjusting the value of $\alpha$, the relative importance of knowledge transfer from the teacher network and the direct supervision of the student network can be controlled during the training process \cite{KD}.
\subsection{Producing Adversarial Examples}
Projected Gradient Descent (PGD) is an iterative method used to generate adversarial examples. This process begins at a randomly selected location (within the \(\epsilon\)-ball), starting from an original input \(x\), and perturbs the image in the direction of maximum loss (Eqn. (\ref{eq:PGD})). The magnitude of these perturbations is governed by a predefined step size, denoted as \(\alpha\), and the process is repeated for a specified number of steps. At each iteration, PGD ensures that the perturbations do not exceed the predefined size, \(\epsilon\), by projecting the perturbed input back into the valid space defined by the \(\epsilon\)-ball~\cite{PGD}. Additionally, Mandry et al.~\cite{PGD} propose that training classifiers with adversarial examples crafted by PGD can enhance their robustness against various types of first-order attacks.
\begin{align}
x^{(t+1)} = \text{Project}_\epsilon \left( x^{(t)} + \alpha \cdot \text{sign} \left( \nabla_x \mathcal{L}(\theta, x, y) \right) \right)
\label{eq:PGD}
\end{align}
Adversarial examples are known for their property of transferability, wherein an example crafted for a specific network often tends to be misclassified by another network\cite{ExHAE}. This indicates that a model is likely to transfer adversarial characteristics effectively across different network architectures. We strategically leverage this transferability characteristic in our approach to diminish the student model's ability to accurately interpret and replicate information derived from the teacher model.
\subsection{Sparse Logits}

The \textit{Stingy Teacher} work suggests that smoothness in the output distribution yields enhancements in the student model's KD performance; the feature that renders a teacher model resistant to distillation is the presence of sparsity in its outputs. This sparsity is achieved by retaining only the top $n$ class probabilities in the logits, effectively focusing the model's attention on the most significant classes while disregarding the less relevant ones. This methodology strategically channels the distillation process, degrades the performance of adversary\cite{ST}.

\begin{figure}[!htb]
  \centering
  \includegraphics[height=4cm]{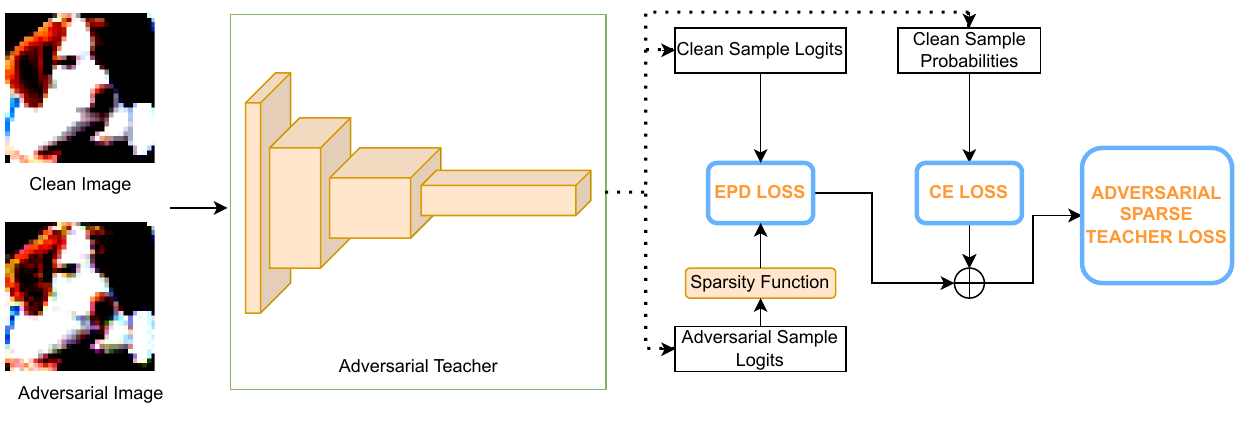}
  \caption{Traning scheme of \textit{AST}. KL loss indicates Kullback-Liebler divergence loss and it uses logits of adversarial and clean samples. CE loss denotes Cross-entropy loss of clean sample probabilities and labels as usual. \textit{AST} loss is sum of these two loss terms.
}
  \label{fig:diagram}
\end{figure}

\subsection{Proposed Method: Adversarial Sparse Teacher}

The primary objective of our study is to develop a teacher model that generates adversarial logits to disrupt the efficacy of models attempting to extract information from these outputs. Therefore, the ultimate design goal is to mislead student models without compromising the performance of the teacher model. To achieve this, we start by generating an adversarial version of the original dataset using a base model, which shares its architectural framework with the \textit{Adversarial Sparse Teacher (AST)}. The training scheme of AST is illustrated in Fig.~\ref{fig:diagram}. The adversarial images generated from the base model, along with the original images, are then used to train our teacher model. Since adversarial examples are crafted to maximize the model's loss, their outputs inherently reflect this characteristic. In this context, we aim to reduce the distance between the outputs of the AST for original and adversarial images, thereby incorporating adversarial logits into the output distribution. Concurrently, our method aims to increase the overall entropy of the output distribution while maintaining the augmented sparseness. Drawing from the foundational concepts of the \textit{Stingy Teacher} approach, our strategy proposes a more robust and secure framework for model training. Unlike the \textit{Stingy Teacher} method, which primarily relies on using the top $n$ logits from the base model's outputs to induce sparsity during the student model’s training—without the need for actively training a teacher model—our approach involves developing a teacher model that generates ambiguous outputs. This is achieved through an optimization process that elevates adversarial sparse logits, utilizing adversarial images. Importantly, this method retains much of the original accuracy of the base model. The key advantage of our approach is its practical application: by training a teacher model to inherently produce modified outputs, we effectively reduce the risk of the teacher model's outputs being compromised or stolen. This not only preserves the integrity of the teacher model but also significantly bolsters its defense against extraction attacks, presenting a more secure and viable solution for real-world applications.

In this context, we defined the loss function of our model as in Eqn.(\ref{eq:example2}).
\begin{align}
 \mathcal{L_{AST}} =
    \alpha \tau^2 \; \mathcal{L_{EPD}}(\sigma_\tau(p_{f_{\theta}}(x^{(i)})), \sigma_\tau(sparse(p_{f_{\theta}}(x^{(i)}_{adv}),\beta))) \nonumber \\ + (1-\alpha)\mathcal{L_{CE}}(\sigma(p_{f_{\theta}}(x^{(i)})), y^{(i)})\label{eq:example2}
\end{align}

Here, the AST network is represented as $f_{\theta}( \cdot )$ with $\theta$ denoting the network's parameters. Training dataset samples alongside their corresponding labels are denoted as $(x^{(i)},y^{(i)})$, and samples' adversarial counterparts are represented as $x^{(i)}_{adv}$. The logit response obtained from either $x^{(i)}$ or $x^{(i)}_{adv}$ by the network $f_{\theta }( \cdot )$ is denoted as $p_{f_{\theta }}(\cdot)$. The $sparse$ function, accepts $p_{f_{\theta }}(\cdot)$ for a given sample and a parameter $\beta$, which represents the sparsity ratio. This function is designed to return the sparse version of the logits. Additionaly, the softmax temperature function is denoted as $\sigma_\tau(\cdot)$. An increase in the $\tau$ parameter, causes this function to yield a distribution characterized by higher entropy, more evenly spread probability across classes. The first term in the equation calculates our proposed \textit{Exponential Predictive Divergence} between the logits of original images and sparse logits of the adversarial examples. Concurrently, the second term computes the cross-entropy loss of training samples and their labels. The goal of this proposed objective function is to minimize the sum of these two terms, thereby enhancing the model's robustness and performance. 

\subsection{Exponential Predictive Divergence (EPD) Loss}

In the proposed method, we observed that using KL divergence loss constrained our ability to adjust these logits as we desired. We sought for obtaining teacher logits that have high entropy with some adversarial peaks for confusing students. To achieve that, we propose a novel loss function, termed exponential predictive divergence (EPD). The EPD function modulates the predictions of the trained model by applying a weight proportional to the target value. Specifically, considering a given target probability, it rapidly increases the model's predictions when there is a deviation from a high target probability. For low target probabilities, including values as low as zero, it gently increases predictions with a slow rate. This gradual adjustment increases the entropy of the trained teacher's predictions, which makes model emulation or 'stealing' more difficult for a student model due to the increased ambiguity in the teacher outputs. The EPD function is provided in Eqn. (\ref{eq:proposed_loss}).

\begin{align}
D_{EPD} (T, P) = \sum_{i} e^{T(i)} \cdot (T(i) - P(i))
\label{eq:proposed_loss}
\end{align}

In this equation, the divergence function takes two probability distributions as input: \(P\), representing the predicted distribution to be updated, and \(T\), representing the target distribution. \(T(i)\) and \(P(i)\) are the logits of the corresponding distributions. The exponential component \(e^{T(i)}\) adjusts these differences based on the value of \(T(i)\). This function weights the differences proportionately to the target value, derived from the teacher model's response to the adversarial example, to refine the model's predictions. Unlike KL divergence, which often results in sharply sparse outputs, our proposed function blends sparse probabilities into the overall distribution (Fig.~\ref{fig:logitsameclass}). This approach keeps the output entropy high, hence confusing the student during distillation process while preserving the accuracy of the teacher model. We note that the EPD function is also advantageous for handling numerical issues associated with small probabilities, which is a common challenge when using the KL divergence function in optimization.


\section{Experimental Settings}
We conducted our experiments using two prominent datasets in this field: CIFAR-10 and CIFAR-100. For the CIFAR-10 dataset, we employed a ResNet18 model as the teacher network. A diverse range of student network architectures was tested, including a basic 5-layer CNN, ResNet18 and two simplified ResNet architectures specifically tailored for CIFAR-10; ResNetC-20, ResNetC-32 by He et al. \cite{he2016deep}. In experiments with the CIFAR-100 dataset, we explored the effects of network capacity and dataset complexity using ResNet18 and ResNet50 as teacher networks. For student networks in these experiments, we utilized ShuffleNetV2 \cite{ma2018shufflenet}, ResNet18 and the respective teacher network architectures themselves. Our experimental setup, including configurations and parameters, closely followed those outlined in the \textit{Nasty Teacher} study for a comprehensive comparison \cite{NT}.
Specifically, the distillation temperatures (denoted as $\tau$) for CIFAR-10 and CIFAR-100 were set at 4 and 20, respectively, in accordance with the recommendations from \cite{NT}.
The weighting factor $\alpha$ was selected as 0.04 for CIFAR-10 and 0.005 for CIFAR-100. For the distillation process, the weighting factor was set to 0.9 for all experiments. Training of the basic CNN was carried out over 100 epochs with a learning rate of 0.001, using the Adam optimizer \cite{kingma2014adam}. Other network architectures were trained using the SGD optimizer, incorporating a momentum of 0.9 and a weight decay of 0.0005. These networks are trained with an initial learning rate of 0.1, adjusting for CIFAR-10 over 160 epochs with reductions in the learning rate by a factor of 10 at the 80th and 120th epochs. For CIFAR-100, the training spanned 200 epochs, with learning rate adjustments by a factor of 5 at the 60th, 120th, and 160th epochs, adhering strictly to the protocol established in the \textit{Nasty Teacher} publication \cite{NT}.

In \textit{Stingy Teacher} experiments, we closely follow the settings in the related paper \cite{ST}. In this context, the sparsity ratios for these models are defined as 0.2 for CIFAR-10 and 0.1 for CIFAR-100. 

The training procedure for our models involves distinct configurations based on the dataset and model architecture. For the ResNet18 \textit{AST} model trained on the CIFAR-10 dataset, the setup includes a temperature parameter ($\tau$) of 6, an $\alpha$ parameter of 0.05, and a sparsity ratio for logits of 0.2. When adapting this ResNet18 \textit{AST} model for the CIFAR-100 dataset, adjustments are made with the temperature parameter increased to 20, the $\alpha$ parameter set to 0.03, and the sparsity ratio of logits fine-tuned to 0.02. Furthermore, the training of the ResNet50 \textit{AST} model on the CIFAR-100 dataset maintains a temperature parameter of 20 and the $\alpha$ of 0.03, and the sparsity ratio of logits is slightly increased and set to 0.03.

Projected Gradient Descent (PGD) attack\cite{PGD}, is implemented on images with the configuration settings as follows: the $\epsilon$ parameter, defining the maximum perturbation size, is established at 0.3; the step size for each iteration is determined to be 0.01; and the total number of iterative steps is fixed at 40.



\begin{table}[tb]
    \caption{Model accuracies obtained using CIFAR-10 dataset. In the Teacher Type column, the following abbreviations are used: ST for \textit{Stingy Teacher}, BASE for Baseline Teacher, NT for \textit{Nasty Teacher} and AST, our proposed model. Best performances are highlighted with bold colors, and second best values are underlined for each student model.
}
    \label{tab:table1}
  \centering
  \begin{tabular}{@{}lllllll@{}}
    \hline
 \textbf{Teacher}&\textbf{Model}  & \textbf{Teacher} & \multicolumn{4}{c}{\textbf{Student Accuracy $\downarrow$}} \\
\cline{4-7}
  \textbf{Type} & & \textbf{Acc.\% $\uparrow$} & \textbf{CNN} &\textbf{ResNetC20} & \textbf{ResNetC32} &\textbf{ResNet18} \\
\hline
Std. & -- & -- & 86.64 & 92.37 & 93.41 & 95.03 \\
\hline
\hline
ST & Res18 & 95.03 & 83.11(-3.53) & \textbf{67.98}(-24.39) & \textbf{74.08}(-19.33) & \textbf{92.47}(-2.56) \\
\hline

BASE & Res18 & \textbf{95.03} & 87.76(+1.12) & 92.27(-0.1) & 92.94(-0.47) & 95.39(+0.36) \\

NT & Res18 & 94.37(-0.66) & 82.98(-3.66) & 88.54(-3.83) & 90.07(-3.34) & 93.76(-1.27) \\

AST (Ours)& Res18 & 94.61(-0.42) & \textbf{79.82}(-6.82) & \underline{87.08}(-5.29) & \underline{88.70}(-4.71) & \underline{93.66}(-1.37)\\
\hline

  \end{tabular}
\end{table}

\section{Results and Discussions}
\subsection{CIFAR-10 Experiments}
Experimental results on CIFAR-10 are presented in Table \ref{tab:table1}. The Std. row (first row) depicts the baseline performances of the student networks. 
In the Teacher accuracy column we want to obtain accuracy as close to Base model as possible. The experiments with CIFAR-10 shows that AST (teacher) model performance is comparable with the other teacher models. In this context, NT and AST models are performing similarly but AST is slightly superior than NT. However, we observe better performance (considerably lower scores) on the students that are distilled from AST. The Table shows the relative reductions in the student performances next to each score in parenthesis. 
ResNetC-20 and ResNetC-32 students show a minor accuracy drop of approximately 0.1\% and 0.47\% post KD from their baseline teacher performances, respectively. Other student architectures report improvements up to around 1\%. 
In the comparison of KD effectiveness among NT and AST, which are similar in that they both train an undistillable teacher model to prevent knowledge transfer, both NT and AST lead to reduced student performance, particularly in simpler networks. However, AST surpasses NT in all students. 

Note that, \textit{Stingy Teacher}(ST) model employs a different methodology, namely the black-box protection, where only a subset of baseline teacher's logits is revealed to the students without explicitly training a teacher model. So, in this model teacher model's outputs are not perturbed, but filtered according to their values. This approach significantly impacts smaller ResNet models, surpassing AST's performance. Nonetheless, for basic CNN student, to our surprise, AST could also outperform ST.
\begin{table}[tb]
    \caption{Model accuracies obtained using CIFAR-100 dataset. In the Teacher Type column, the following abbreviations are used: ST for \textit{Stingy Teacher}, BASE for Baseline Teacher, NT for \textit{Nasty Teacher} and AST, our proposed model, for \textit{AST}. Best performances are highlighted with bold colors, and second best values are underlined for each student model.}
    \label{tab:table2}
  \centering
  \begin{tabular}{@{}llllll@{}}
    \hline
    
  \textbf{Teacher} & \textbf{Model} & \textbf{Teacher} & \multicolumn{3}{c}{\textbf{Student Accuracy$\downarrow$
}} \\
\cline{4-6}
\textbf{Type} &  & \textbf{Acc.\% $\uparrow$
} & \textbf{ShuffleNetV2} & \textbf{ResNet18} & \textbf{Teacher 's Arch.} \\  
\hline
Std. & -- & -- & 72.10  & 78.28 & -- \\

\hline
\hline
ST & Res18 & 78.28& \underline{50.49}(-21.61)  & \underline{55.30}(-22.98) & \underline{55.30}(-22.98)\\
\hline
BASE & Res18 & \textbf{78.28} & 74.38(+2.28)  & 79.12(+0.84) & 79.12(+0.84) \\
NT & Res18 & 77.80(-0.48) & 65.01(-7.09)  & 74.68(-3.60) & 74.68(-3.60) \\
AST (Ours) & Res18 & 77.02(-1.26) & \textbf{4.37}(-67.73) & \textbf{44.01}(-34.27) & \textbf{44.01}(-34.27) \\
\hline
\hline
ST & Res50 & 77.55& \underline{46.46}(-25.64) & \textbf{54.22}(-24.06) & \underline{54.14}(-23.41)\\
\hline
BASE & Res50 & 77.55 & 74.00(+1.90) &  79.27(+0.99) & 80.03(+2.48) \\
NT & Res50 & 76.88(-0.67) & 67.14(-4.96)  & 73.87(-4.41) & 75.99(-1.56) \\
AST (Ours) & Res50 & \textbf{77.69}(+0.14)& \textbf{26.32}(-45.78) & \underline{58.63}(-19.65) & \textbf{46.62}(-30.93) \\
\hline
  \end{tabular}
\end{table}

\subsection{CIFAR-100 Experiments}
The experimental results on CIFAR-100 are provided in Table \ref{tab:table2}. In this setting we conducted experiments using two different teacher architectures, ResNet18 and ResNet50. Since there are more categories in this dataset, we want to observe the effect of capacity increase to the KD performances. The results show that with the ResNet18 architecture, AST (teacher) model exhibits slightly lower performance compared to the NT model. Conversely, when employing the more complex ResNet50 architecture, AST surpasses all teacher models, including the baseline teacher.

Our proposed model demonstrates that in a fully transparent (full white-box) setting, i.e. when we compare AST with the NT, AST exhibits a more pronounced accuracy decline. In other words, it outperforms the NT method, which is a fully trained teacher model for IP protection. Specifically, the defensive performance of AST increases significantly for ShuffleNetV2 student models for both teacher architectures; around 67.73\% decrease is observed in ResNet18 and 45.78\% in ResNet50 AST models compared to the baseline student model. The performance of AST is also superior than the ST model for almost all students except ResNet18 student distilled from ResNet50 teacher. 

\subsection{Data-free and Subset Experiments}

In order to emulate a more realistic scenario, we conducted experiments using smaller training set sizes for both CIFAR-10 and CIFAR-100 datasets. Typically, attackers attempting to steal a model do not have access to the full dataset. This is particularly important in fields like the medical domain, where data is both limited and sensitive. Ensuring robust defenses in these scenarios helps protect valuable and restricted datasets from being exploited. Therefore, in this section, we analyzed the impact of a reduced labeled dataset, which mainly reflects the stealing effort from the teacher's logit responses, without substantial support from the labeled dataset during distillation. In this context, we utilized subsets containing 50, 100, and 250 images per class, as well as the full dataset, which consists of 500 images per class for CIFAR-100 and 5000 images per class for CIFAR-10. By comparing the results across different subsets, we aim to highlight the effectiveness of our defense method in protecting models trained with limited data. The results of our experiments are summarized in Table \ref{tab:subsetexcf10} for CIFAR-10 and Table \ref{tab:subsetex} and for CIFAR-100.


\begin{table}[!htb]
    \caption{Results of subset experiments conducted on the CIFAR-10 dataset using ResNet18 architecture for both the teacher and student models.} 
    \label{tab:subsetexcf10}
  \centering
  \begin{tabular}{@{}lllll@{}}
    \hline
\textbf{Model Type} & \textbf{Images per class}& \textbf{Base Acc.\%} & \textbf{NT Acc.\%} & \textbf{AST Acc. \%} \\  
\hline
Teacher&--&95.03&94.37&94.61\\
Student&50&39.30(-55.73)&31.91(-62.46)&\textbf{17.52}(-77.09)\\
Student&100&49.021(-46.01)&40.29(-54.08)&\textbf{31.54}(-63.07)\\
Student&250&61.58(-33.45)&58.49(-35.89)&\textbf{47.20}(-47.41)\\
Student&5000&95.39(+0.36)&93.76(-0.61)&\textbf{93.66}(-0.95)\\

\hline
  \end{tabular}
\end{table}

\begin{table}[!htb]
    \caption{Results of subset experiments conducted on the CIFAR-100 dataset using ResNet18 architecture for both the teacher and student models. }
    \label{tab:subsetex}
  \centering
  \begin{tabular}{@{}lllll@{}}
    \hline
\textbf{Model Type} & \textbf{Images per Class}& \textbf{Base Acc.\%} & \textbf{NT Acc.\%} & \textbf{AST Acc. \%} \\  
\hline
Teacher&--&78.28&77.80&77.02\\
Student&50&54.92(-23.38)&26.74(-51.06)&\textbf{2.34}(-74.68)\\
Student&100&69.35(-8.93)&48.28(-29.52)&\textbf{5.49}(-71.53)\\
Student&250&76.79(-1.49)&67.42(-10.38)&\textbf{18.20}(-58.82)\\
Student&500&79.12(+0.84)&74.31(-3.49)&\textbf{44.01}(-33.01)\\

\hline
  \end{tabular}
\end{table}

In the presence of limited data, AST significantly lowers the performance of the student model and outperforms NT. In the case of utilizing 50 labeled images, the distilled student performance has dropped 77.09\% in CIFAR-10 dataset, 74.68\% in CIFAR-100 dataset. We can see from the tables that the protection in the case of low data regime is significantly higher than NT model.


We also conducted an experiment to simulate a scenario where the adversary has access to some unlabeled data but no labeled data. In this setting, we removed the Cross Entropy term from the total loss calculation during KD since there is no labeled data involved. The combined results of this experiment for both datasets are shown in Table \ref{tab:combined}. In these experiments, the full training data is utilized without labels (see the "KD w/o labels" row). As evident from the table, the perturbation to the teacher logits introduced by the AST method severely disrupts the performance of the student, while the teacher's performance remains high. From these results, we observe that the performance gap between NT and AST narrows as the amount of labeled data used during distillation increases. However, this is typically not the case in practice, especially in domains where data is valuable and generating labeled data is expensive, making model stealing attempts more challenging. Therefore, for IP protection in such domains, AST provides better protection.

\begin{table}[!htb]
    \caption{Results of the unlabeled data experiments for CIFAR-100 and CIFAR-10 using ResNet18. "KD w/ labels" row shows the student performance trained using labeled data and traditional KD loss. "KD w/o labels" row shows the student performance trained using unlabeled data and KD loss without the Cross Entropy term.}
    \label{tab:combined}
    \centering
    \begin{tabular}{@{}lcccccc@{}}
        \toprule
        & \multicolumn{3}{c}{\textbf{CIFAR-100}} & \multicolumn{3}{c}{\textbf{CIFAR-10}} \\
        \cmidrule(lr){2-4} \cmidrule(lr){5-7}
        & \textbf{BT Acc.\%} & \textbf{NT Acc.\%} & \textbf{AST Acc.\%} & \textbf{BT Acc.\%} & \textbf{NT Acc.\%} & \textbf{AST Acc.\%} \\
        \midrule
        Teacher & 78.28 & 77.80 & 77.02 & 95.03 & 94.37 & 94.61 \\
        KD w/ labels & 79.12 & 74.68(-3.12) & 44.01(-33.01) & 95.39 & 93.76(-0.61) & 93.66(-0.95) \\
        KD w/o labels & 79.32 & 73.60(-4.20) & \textbf{28.38}(-48.64) & 95.38 & 94.09(-0.28) & \textbf{92.97}(-1.64) \\
        \bottomrule
    \end{tabular}
\end{table}

The results show that NT maintains similar performance when labels are not available. However, our model significantly reduces the student performance and outperforms NT.

\subsection{Qualitative and Quantitative Analysis}
\label{sec:qualitative}
\begin{figure}[!htb]
  \centering
  \includegraphics[height=8cm]{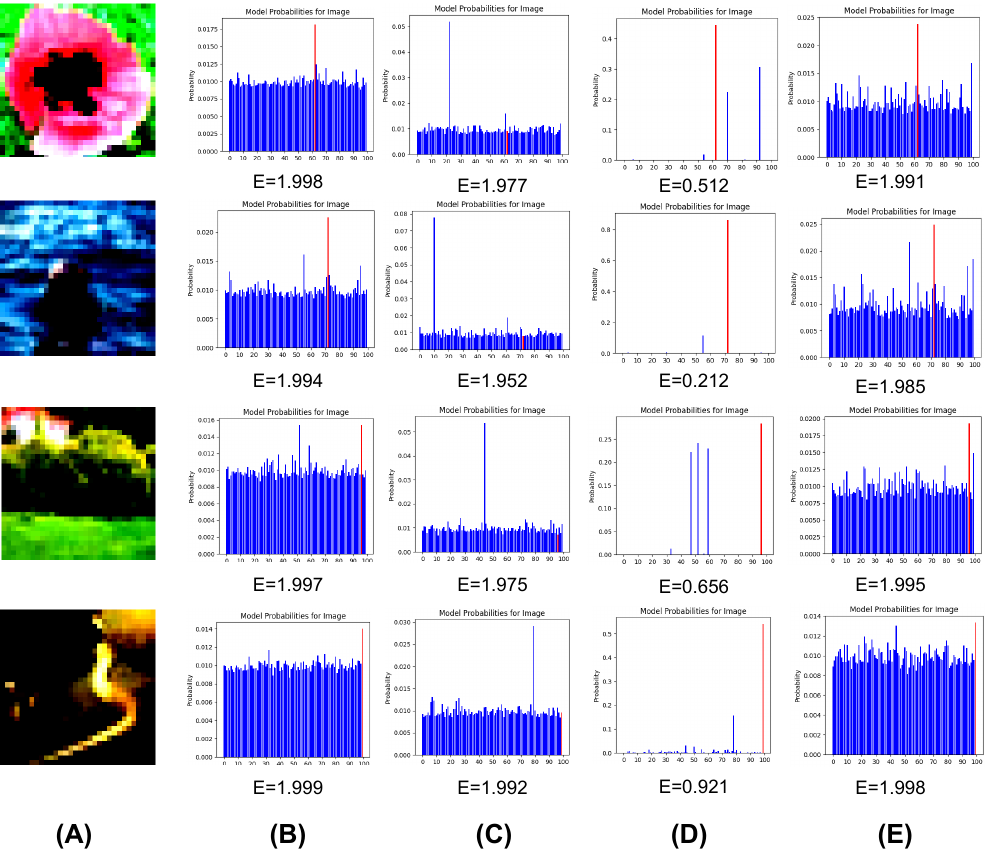}
  \caption{Illustration of the logit responses after the application of softmax temperature. Columns depict the following information: (A) samples of clean images,  (B) responses of the baseline model to these clean samples, (C) responses of the baseline model to the adversarial counterparts of the clean images, (D) responses of the NT model to clean samples, (E) responses of our proposed method (AST) to clean images. Below each distribution, the entropy of the distributions are provided.}
  \label{fig:logits}
\end{figure}

This section complements the previous experiment results with additional qualitative and quanitative analysis. Given that the CIFAR-100 dataset contains ten times more categories than the CIFAR-10 dataset, our focus will be on demonstrating the ResNet18 model's responses specifically for CIFAR-100. Output responses of different networks focusing on the baseline teacher, NT and AST models, are illustrated in Fig.~\ref{fig:logits}.  Similar to the approach in \cite{ST}, we applied a softmax temperature to the logit responses to enhance visualization in this paper. Additionally, we present the output responses of the baseline teacher model to corresponding adversarial images (shown in the third column), and the entropy values of each distribution are displayed beneath them. The baseline model typically yields a nearly uniform distribution across classes, with a single peak in a class, demonstrating high confidence for both clean and adversarial images; frequently produce peaks for the correct class in the clean images and an incorrect class in the adversarial images. The distribution for adversarial image responses exhibits lower entropy compared to that for clean images. Contrary to these distributions, the distribution from the AST model reveals additional, sparse peaks. We attribute these peaks to our sparse optimization scheme (with EPD loss function), which generally does not exceed the true category in magnitude. We believe that the increased \sout{relative} entropy, coupled with the additional sparse peaks, renders the distillation process more challenging for the student models. 

\begin{figure}[!htb]
  \centering
  \includegraphics[height=8cm]{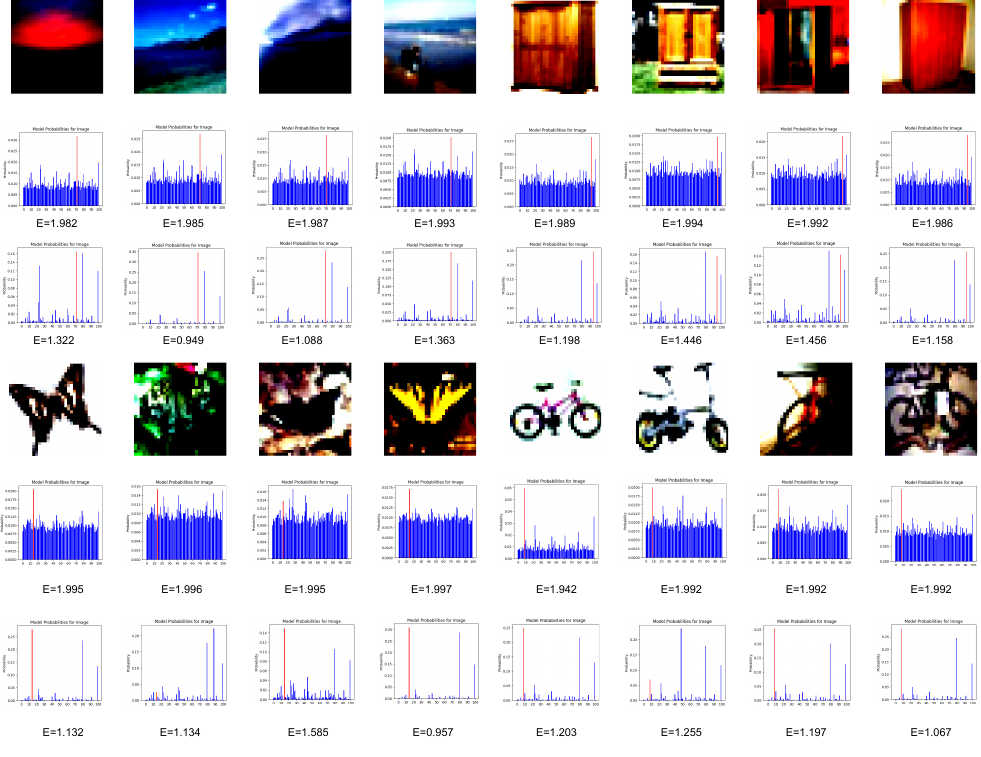}
  \caption{The responses of our ResNet18 AST model and AST trained with KL Divergence model to clean samples from two sets of identical classes is displayed. Top row: input images, middle-row:Responses of AST trained with EPD loss, last row: responses of AST trained with KL Divergence loss. Samples in the leftmost 4 columns and righmost 4 columns belong to same classes. The softmax temperature applied to all logits and below each distribution, the entropy value is displayed.}
  \label{fig:logitsameclass}
\end{figure}

In addition, we want to compare the AST responses trained with EPD loss and KL divergence loss within a specific class to examine our model's general characteristics. In this context, Fig.~\ref{fig:logitsameclass} presents the models' responses for randomly selected samples within the same category for four different categories. As observed in the figure, AST (with EPD) model's response characteristics are notably similar, characterized by high entropy (a uniform-like distribution across all categories), yet with sparse and relatively consistent peaks for a few classes (approximately 6-8). Notably, a higher peak is evident in the true category, which is highlighted in red. AST (with KL-div) responses are more sparse and has lower entropy. After the optimization, the high amount of sparseness inversely effects (degrades) the teacher performance.

Furthermore, in order to quantify the characteristics of the model responses, we computed the entropies of the output distributions for the entire test set for all teacher models. The statistics of the entropy values are provided for two categories in Table \ref{tab:within_class_ent_kl}. The quantitative results are on par with the qualitative results and consistent in different categories, where base teacher and AST output entropies are higher compared to NT model. However, AST entropy variance is higher than the base teacher model, due to relatively peaky nature of the output distributions arising due to the sparseness constraint applied during training.  

\begin{table}[tb]
    \caption{Within Class KL Diverences and Entropies of ResNet18 Model Outputs in CIFAR-100 for two categories.}
    \label{tab:within_class_ent_kl}
  \centering
  \begin{tabular}{@{}lllllllllll@{}}
    \hline
& & \multicolumn{4}{c}{\textbf{KL Divergence}}& \multicolumn{4}{c}{\textbf{Entropy}} \\
\cline{3-6}
\cline{8-11}
\textbf{Model}&\textbf{Label}&\textbf{Mean}&\textbf{Variance}&\textbf{Min}&\textbf{Max}&&\textbf{Mean}&\textbf{Variance}&\textbf{Min}&\textbf{Max}\\
\hline
\centering
BASE & 0 & 0.00082 & 8.56e-8 & 0 & 0.00215 &&1.99831 & 4.63e-8 & 1.99626 & 1.99958\\
NT & 0 & 0.49432 & 0.26566 & 0 & 3.63332 &&0.25185 & 0.14789 & 0.00895 & 1.68001 \\
AST & 0 & 0.00240 & 3.68e-6 & 0 & 0.01016 &&1.98986 & 1.55e-5 & 1.98219 & 1.99746\\
\hline
BASE & 10 & 0.00161 & 2.61e-7 & 0 & 0.00286&& 1.99841 & 8.11e-7 & 1.99361 & 1.99954  \\
NT & 10 & 1.06778 & 1.20049 & 0 & 4.21843 && 0.70504 & 0.21716 & 0.00968 & 1.74953\\
AST & 10 & 0.01149 & 4.42e-5 & 0 & 0.04374&& 1.98177 & 0.00058 & 1.87968 & 1.99830 \\

\hline
  \end{tabular}
\end{table}

We further investigated the consistency of output distributions within the same classes by designing another set of measurements that involves randomly selecting a reference sample from two categories and calculating its relative entropy against the outputs of other samples in the same category. The findings, detailed in Table \ref{tab:within_class_ent_kl}, indicate significant variability among the models. The base model consistently shows the lowest mean KL divergence for both labels, as expected, which suggests a closer correlation with the expected distribution, denoting confident and precise model predictions. This is a desired property for recognition, yet not desirable for protection. In contrast, the NT model exhibits greater variance in KL divergence, particularly for label 10, reflecting a wider deviation from the standard distribution and potentially indicating less consistent predictions designed to mislead the student model. In comparison, the AST model achieves a lower KL divergence relative to the NT model, yet higher than the base model, underscoring the stability of its output signals. Moreover, it presents a significantly greater variance in divergence values compared to the base model. Based on the entropy and KL divergence values, we can conclude that the desired outputs are being generated for misleading the student models.

\subsection{Ablation Studies}


In this section, we examine the effects of various parameters on the performance and defensiveness of the ResNet18 architecture trained on the CIFAR-100 dataset. Specifically, we examine the influence of the sparsity ratio, the weight parameter \(\alpha\), and the temperature parameter \(\tau\). These parameters play crucial roles in balancing teacher performance and defensiveness. Additionally, we explore the impact of different loss functions, including our proposed EPD loss, to understand its effect on the behavior of the AST model.

The first experiment focuses on the parameters \(\alpha\) and \(\tau\). These parameters function as weighting factors on the total loss due to the term \(\alpha\tau^2\). Additionally, the temperature parameter \(\tau\) scales the logits for both clean and adversarial logits in the softmax function by dividing the logits by \(\tau\). In the experiments, three different \(\alpha\) values (0.03, 0.035, and 0.04) are used to observe how both lower and higher values affect the total loss. Additionally, four different \(\tau\) values (10, 20, 30, and 40) are employed for the same reason and to examine the effect of scaling the logits. This experiment is conducted with two different settings of sparsity ratios: 0.02 and 0.03. The results of the experiments using a sparsity ratio of 0.02 are presented in Table \ref{tab:absr02}, and the results using a sparsity ratio of 0.03 are presented in Table \ref{tab:absr03}. These sparsity ratios were chosen because lower sparsity ratios have been shown to have a more favorable impact on performance, as will be explained in detail later in this section.

In general, for both sparsity ratio choices, teacher performance increased with a smaller weight parameter (\(\alpha\)). This is due to the decreased effect of the EPD loss and the increased influence of Cross Entropy. However, reducing this parameter did not necessarily result in increased defensiveness. To achieve the best teacher and defense performance simultaneously, the \(\tau\) and sparsity ratio factors play a major role, and the effects of all parameters should be considered together.

Setting the temperature parameter (\(\tau\)) between 20 and 30 generally increased defensiveness and resulted in a notable decrease in student accuracy while keeping teacher accuracy high. On the other hand, smaller values of \(\tau\) often enhanced teacher performance but decreased defensiveness due to the lowered effect of EPD loss. Conversely, higher values of \(\tau\) led to a decline in both defensiveness and teacher performance. This is because setting the \(\tau\) parameter too high reduces the impact of the correct class, consequently decreasing teacher performance. Additionally, higher temperature values lower the distance between adversarial logits and remaining logits, thereby reducing the effect of adversarial logits and causing decreased defensiveness. Temperature ($\tau$) value of 20 and sparsity ratio of 0.03 allowed us to attain ideal performance, which resulted in a 33.01\% decrease in student accuracy. Because the term $\alpha\tau^2$ is used to determine the total loss, there is a strong dependence between the $\alpha$ and $\tau$ parameters. Thus, both teacher performance and defensiveness are taken into consideration for choosing these parameters.
\begin{table}[!htb]
    \caption{The table shows the effect of varying values of the $\alpha$ and $\tau$ parameters on a Resnet-18 \textit{AST} network that was trained using a sparsity ratio of 0.02 using the CIFAR-100 dataset.}
    \label{tab:absr02}
  \centering
      \begin{adjustbox}{width=\textwidth*8/9}

\begin{tabular}{@{}lllllllllllllll@{}}
\hline
\textbf{$\alpha$}&\multicolumn{4}{c}{\textbf{0.03}}&&\multicolumn{4}{c}{\textbf{0.035}}&&\multicolumn{4}{c}{\textbf{0.04}} \\
\cline{2-5}
\cline{7-10}
\cline{12-15}
\textbf{$\tau$}&\textbf{10}&\textbf{20}&\textbf{30}&\textbf{40}&&\textbf{10}&\textbf{20}&\textbf{30}&\textbf{40}&&\textbf{10}&\textbf{20}&\textbf{30}&\textbf{40}\\
\hline
\textbf{Teacher}&\textbf{77.87}&77.02&76.80&75.69&&77.21&76.53&76.13&75.27&&76.62&76.67&75.28&74.57\\
\textbf{Student}&78.55&\textbf{44.01}&57.24&74.53&&78.40&52.46&60.87&72.65&&50.91&64.05&71.73&76.30\\

\hline
  \end{tabular}
            \end{adjustbox}

\end{table}

\begin{table}[!htb]
    \caption{The table shows the effect of varying values of the $\alpha$ and $\tau$ parameters on a Resnet-18 \textit{AST} network that was trained using a sparsity ratio of 0.03 using the CIFAR-100 dataset.}
    \label{tab:absr03}
  \centering
    \begin{adjustbox}{width=\textwidth*8/9}

\begin{tabular}{@{}lllllllllllllll@{}}
\hline
\textbf{$\alpha$}&\multicolumn{4}{c}{\textbf{0.03}}&&\multicolumn{4}{c}{\textbf{0.035}}&&\multicolumn{4}{c}{\textbf{0.04}} \\
\cline{2-5}
\cline{7-10}
\cline{12-15}
\textbf{$\tau$}&\textbf{10}&\textbf{20}&\textbf{30}&\textbf{40}&&\textbf{10}&\textbf{20}&\textbf{30}&\textbf{40}&&\textbf{10}&\textbf{20}&\textbf{30}&\textbf{40}\\
\hline
\textbf{Teacher}&\textbf{77.60}&76.65&76.75&75.83&&77.16&76.38&75.99&75.43&&77.20&76.38&75.45&74.88\\
\textbf{Student}&78.84&72.50&56.47&63.60&&78.32&\textbf{54.96}&59.15&73.09&&75.87&56.46&71.71&69.51\\

\hline
  \end{tabular}
            \end{adjustbox}

\end{table}

The second ablation was conducted to assess the impact of the sparsity ratio on the model's performance. The sparsity ratio determines how many adversarial logits are used and affects the value of these logits due to the softmax function applied to them. A lower sparsity ratio means each adversarial logit has a higher value compared to a higher sparsity ratio.

In this context, two best potential parameter set from the previous ablation study is utilized to explore more on the sparsity ratio parameter from Table \ref{tab:absr02} and \ref{tab:absr03}, considering the settings where student performances are lowest. Hence, we varied the sparsity ratio for $\alpha$=0.03 and $\tau$=20 and obtained the results in Table \ref{tab:table5_3}. In addition, we varied the sparsity ratio for $\alpha$=0.035 and $\tau$=20 and obtained the results in Table \ref{tab:table5_2}. In both cases, the best teacher performance is noted when sparsity is not used, which implies that the defense mechanism will be eliminated. Increased defense is typically the result of lower sparsity ratios, except when using only one sparse class, which generally means using only the true label. Setting a low sparsity ratio effectively increases the value of sparse classes due to the softmax function applied to these logits. This increased value amplifies the impact of these logits, leading to better defensiveness, especially because of the exponential term in the EPD function.

\begin{table}[!htb]
    \caption{The table shows how different sparsity ratios affect a ResNet18 \textit{AST} model that was trained using the CIFAR-100 dataset and had $\tau$ and $\alpha$ values of 20 and 0.03, respectively.}
    \label{tab:table5_3}
  \centering
    \begin{adjustbox}{width=\textwidth*7/9}

  \begin{tabular}{@{}lllllll@{}}
\hline
\textbf{Teacher}&\textbf{ }&\textbf{Teacher} &\textbf{Student}&\textbf{Teacher-Student}&\textbf{Teacher}&\textbf{} \\
\textbf{Type}&\textbf{SR}&\textbf{Perf.\% $\uparrow$}&\textbf{Perf.\% $\downarrow$}&\textbf{Difference(D)}&\textbf{Difference(P)}&\textbf{D/P}$\uparrow$\\
\hline
BASE&---&78.28&---&---&---&---\\
\hline
AST&0.01&77.07&75.56&1.51&1.21&1.25\\
AST&0.02&77.02&\textbf{44.01}&33.01&1.26&\textbf{26.20}\\
AST&0.03&76.65&72.50&4.15&1.63&2.55\\
AST&0.05&77.40&77.08&0.32&0.88&0.36\\
AST&0.07&77.29&77.86&-0.57&0.99&-0.58\\
AST&0.1&77.12&78.26&-1.14&1.16&-0.98\\
AST&1&\textbf{77.61}&79.05&-1.44&0.67&-2.15\\
\hline
  \end{tabular}
          \end{adjustbox}

\end{table}

\begin{table}[!htb]
    \caption{The table shows how different sparsity ratios affect a ResNet18 \textit{AST} model that was trained using the CIFAR-100 dataset and had $\tau$ and $\alpha$ values of 20 and 0.035, respectively.}
    \label{tab:table5_2}
  \centering
    \begin{adjustbox}{width=\textwidth*7/9}

  \begin{tabular}{@{}lllllll@{}}
\hline
\textbf{Teacher}&\textbf{ }&\textbf{Teacher} &\textbf{Student}&\textbf{Teacher-Student}&\textbf{Teacher}&\textbf{} \\
\textbf{Type}&\textbf{SR}&\textbf{Perf.\% $\uparrow$}&\textbf{Perf.\% $\downarrow$}&\textbf{Difference(D)}&\textbf{Difference(P)}&\textbf{D/P}$\uparrow$\\
\hline
BASE&---&78.28&---&---&---&---\\
\hline
AST&0.01&76.71&70.99&5.72&1.57&3.64\\
AST&0.02&76.53&\textbf{52.46}&24.07&1.75&\textbf{13.75}\\
AST&0.03&76.38&54.96&21.42&1.90&11.27\\
AST&0.05&76.64&62.32&14.32&1.64&8.73\\
AST&0.07&76.94&69.53&7.41&1.34&5.53\\
AST&0.1&76.75&78.25&-1.50&1.53&-0.98\\
AST&1&\textbf{78.24}&79.36&-1.12&0.04&-28.00\\

\hline
  \end{tabular}
          \end{adjustbox}

\end{table}

In order to select the best parameters from this two-value optimization setting (i.e. teacher and student performances), we defined a new metric taking into account defensiveness (D) and performance degradation (P) of the trained model using teacher and student accuracies. In this method, we first determined the model's defensiveness by subtracting the student model's accuracy from the teacher model's. This represents the trained model's capacity for defensiveness. We also computed the difference between the baseline and AST teacher models in order to evaluate the teacher model's performance decline. Our objective is to make the AST model more protective without significantly compromising its performance. We can obtain the optimum values, which maximize defensiveness (D) and minimize the performance degradation (P), by dividing D by P. The model with the highest D/P value is the one we would prefer. As presented in the Tables \ref{tab:table5_3} and \ref{tab:table5_2}, with a D/P value of 26.20, the optimal result is obtained when  $\tau$ is set to 20 and $\alpha$ is set to 0.03, and the sparsity ratio is set to 0.02.

\begin{table}[!htb]
    \caption{The table illustrates the effect of various loss functions on a ResNet18 \textit{AST} network trained with CIFAR-100.}
    \label{tab:table6}
  \centering
  \begin{tabular}{@{}lll@{}}
\hline
\textbf{Loss Function }&\textbf{Teacher} &\textbf{Student} \\
\textbf{}&\textbf{Acc.\%$\uparrow$}&\textbf{Acc.\%$\downarrow$}\\
\hline
KL-Div Loss&67.25&62.34\\
EPD Loss&\textbf{77.02}&\textbf{44.01}\\

\hline
  \end{tabular}
\end{table}

As the final ablation study, we compare AST model performances using KL divergence loss and our proposed EPD loss in Table \ref{tab:table6}. This experiment is conducted to examine the effect of the EPD loss compared to KL divergence loss and to see how adversarial logits work with both loss functions. 
In our experiments, we used two teachers: one trained with KL divergence and the other with EPD loss, both with \(\alpha = 0.03\), \(\tau = 20\), and a sparsity ratio of 0.02. The EPD-trained teacher achieved about 10\% higher accuracy, showcasing the effectiveness of EPD in enhancing performance. 
The KL divergence-trained teacher's lower performance is due to its sparse outputs, which hinder effective knowledge distillation. EPD, on the other hand, produces non-sparse outputs by incorporating adversarial sparse logits while maintaining high entropy.
In terms of defensiveness, the KL divergence-trained teacher caused a 4.9\% drop in student accuracy, whereas the EPD-trained teacher caused a 33.0\% drop, highlighting EPD's superior impact on defense mechanisms.

\section{Conclusion}

In this study, we introduce a new training method for teacher networks called \textit{AST}, designed to safeguard against knowledge theft through Knowledge Distillation. We introduced sparse logits of adversarial images during training of the AST model. Consequently, the AST's responses are deliberately misleading, consistently providing incorrect information to deter adversaries. Our extensive experiments across different teacher-student architectures and datasets demonstrate the effectiveness of our approach. In scenarios where adversaries have complete knowledge, including access to training data, AST significantly impairs their performance. This method is particularly effective in complex teacher architectures and datasets, outperforming other strategies in fully-disclosed model scenarios.  

We also introduce a novel divergence loss function (EPD) utilized in \textit{AST} training that effectively incorporates adversarial logits into the output and has proven effective in our empirical results. However, future research is vital to refine this approach and explore its broader implications, particularly in terms of computational efficiency and adaptability to various model architectures.

\bibliographystyle{ieeetr}
\footnotesize \bibliography{main.bib}

\begin{thebibliography}{10}

\bibitem{KD}
G.~E. Hinton, O.~Vinyals, and J.~Dean, ``Distilling the knowledge in a neural network,'' in {\em NIPS}, 2015.

\bibitem{mirzadeh2020improved}
S.~I. Mirzadeh, M.~Farajtabar, A.~Li, N.~Levine, A.~Matsukawa, and H.~Ghasemzadeh, ``Improved knowledge distillation via teacher assistant,'' in {\em Proceedings of the AAAI conference on artificial intelligence}, vol.~34, pp.~5191--5198, 2020.

\bibitem{romero2015fitnets}
A.~Romero, N.~Ballas, S.~E. Kahou, A.~Chassang, C.~Gatta, and Y.~Bengio, ``Fitnets: Hints for thin deep nets,'' in {\em 3rd International Conference on Learning Representations, {ICLR} 2015, San Diego, CA, USA, May 7-9, 2015, Conference Track Proceedings} (Y.~Bengio and Y.~LeCun, eds.), 2015.

\bibitem{chen2021feature}
H.~Chen, Y.~Wang, C.~Xu, C.~Xu, and D.~Tao, ``Learning student networks via feature embedding,'' {\em {IEEE} Trans. Neural Networks Learn. Syst.}, vol.~32, no.~1, pp.~25--35, 2021.

\bibitem{komodakis2017paying}
N.~Komodakis and S.~Zagoruyko, ``Paying more attention to attention: improving the performance of convolutional neural networks via attention transfer,'' in {\em ICLR}, 2017.

\bibitem{park2019relational}
W.~Park, D.~Kim, Y.~Lu, and M.~Cho, ``Relational knowledge distillation,'' in {\em Proceedings of the IEEE/CVF conference on computer vision and pattern recognition}, pp.~3967--3976, 2019.

\bibitem{passalis2020heterogeneous}
N.~Passalis, M.~Tzelepi, and A.~Tefas, ``Heterogeneous knowledge distillation using information flow modeling,'' in {\em Proceedings of the IEEE/CVF Conference on Computer Vision and Pattern Recognition}, pp.~2339--2348, 2020.

\bibitem{peng2019correlation}
B.~Peng, X.~Jin, J.~Liu, D.~Li, Y.~Wu, Y.~Liu, S.~Zhou, and Z.~Zhang, ``Correlation congruence for knowledge distillation,'' in {\em Proceedings of the IEEE/CVF International Conference on Computer Vision}, pp.~5007--5016, 2019.

\bibitem{gou2021knowledge}
J.~Gou, B.~Yu, S.~J. Maybank, and D.~Tao, ``Knowledge distillation: A survey,'' {\em International Journal of Computer Vision}, vol.~129, pp.~1789--1819, 2021.

\bibitem{SD}
L.~Zhang, J.~Song, A.~Gao, J.~Chen, C.~Bao, and K.~Ma, ``Be your own teacher: Improve the performance of convolutional neural networks via self distillation,'' in {\em 2019 IEEE/CVF International Conference on Computer Vision (ICCV)}, pp.~3712--3721, 2019.

\bibitem{yuan2020revisiting}
L.~Yuan, F.~E. Tay, G.~Li, T.~Wang, and J.~Feng, ``Revisiting knowledge distillation via label smoothing regularization,'' in {\em Proceedings of the IEEE/CVF Conference on Computer Vision and Pattern Recognition}, pp.~3903--3911, 2020.

\bibitem{IP}
M.~Xue, Y.~Zhang, J.~Wang, and W.~Liu, ``Intellectual property protection for deep learning models: Taxonomy, methods, attacks, and evaluations,'' {\em {IEEE} Trans. Artif. Intell.}, vol.~3, no.~6, pp.~908--923, 2022.

\bibitem{adi2018turning}
Y.~Adi, C.~Baum, M.~Cisse, B.~Pinkas, and J.~Keshet, ``Turning your weakness into a strength: Watermarking deep neural networks by backdooring,'' in {\em 27th USENIX Security Symposium (USENIX Security 18)}, pp.~1615--1631, 2018.

\bibitem{uchida2017embedding}
Y.~Uchida, Y.~Nagai, S.~Sakazawa, and S.~Satoh, ``Embedding watermarks into deep neural networks,'' in {\em Proceedings of the 2017 ACM on international conference on multimedia retrieval}, pp.~269--277, 2017.

\bibitem{fan2019rethinking}
L.~Fan, K.~W. Ng, and C.~S. Chan, ``Rethinking deep neural network ownership verification: Embedding passports to defeat ambiguity attacks,'' {\em Advances in neural information processing systems}, vol.~32, 2019.

\bibitem{zhang2020passport}
J.~Zhang, D.~Chen, J.~Liao, W.~Zhang, G.~Hua, and N.~Yu, ``Passport-aware normalization for deep model protection,'' {\em Advances in Neural Information Processing Systems}, vol.~33, pp.~22619--22628, 2020.

\bibitem{NT}
H.~Ma, T.~Chen, T.-K. Hu, C.~You, X.~Xie, and Z.~Wang, ``Undistillable: Making a nasty teacher that cannot teach students,'' in {\em ICLR}, 2021.

\bibitem{KL}
S.~Kullback, {\em Information theory and statistics}.
\newblock Courier Corporation, 1997.

\bibitem{ST}
H.~Ma, Y.~Huang, H.~Tang, C.~You, D.~Kong, and X.~Xie, ``Sparse logits suffice to fail knowledge distillation,'' in {\em ICLR}, 2022.

\bibitem{PGD}
A.~Madry, A.~Makelov, L.~Schmidt, D.~Tsipras, and A.~Vladu, ``Towards deep learning models resistant to adversarial attacks,'' in {\em ICLR}, 2018.

\bibitem{ALP}
H.~Kannan, A.~Kurakin, and I.~J. Goodfellow, ``Adversarial logit pairing,'' {\em CoRR}, vol.~abs/1803.06373, 2018.

\bibitem{ExHAE}
I.~J. Goodfellow, J.~Shlens, and C.~Szegedy, ``Explaining and harnessing adversarial examples,'' in {\em ICLR}, 2015.

\bibitem{he2016deep}
K.~He, X.~Zhang, S.~Ren, and J.~Sun, ``Deep residual learning for image recognition,'' in {\em Proceedings of the IEEE conference on computer vision and pattern recognition}, pp.~770--778, 2016.

\bibitem{ma2018shufflenet}
N.~Ma, X.~Zhang, H.-T. Zheng, and J.~Sun, ``Shufflenet v2: Practical guidelines for efficient cnn architecture design,'' in {\em Proceedings of the European conference on computer vision (ECCV)}, pp.~116--131, 2018.

\bibitem{kingma2014adam}
D.~P. Kingma and J.~Ba, ``Adam: A method for stochastic optimization,'' {\em arXiv preprint arXiv:1412.6980}, 2014.

\end{thebibliography}

\end{document}